\documentclass{article}
\usepackage{spconf,amsmath,amssymb,graphicx}
\usepackage{booktabs}
\usepackage{cite}
\usepackage{url}

\def\L{{\cal L}}

\title{STUNet-Fusion: Spatiotemporal Needle-Tip Localization
in Ultrasound Video via Multi-Channel Motion Fusion}

\name{\textnormal{Chia-Chi Hsu$^{1}$, Chia-Hsuan Hsu$^{2}$, Che-Chou Shen$^{1}$}}

\address{
$^{1}$National Taiwan University of Science and Technology, Taiwan\\
$^{2}$National Yang Ming Chiao Tung University, Taiwan
}

\begin{document}
\maketitle

\begin{abstract}
Needle-tip localization in ultrasound remains challenging because the needle may appear weak, discontinuous, or partially invisible, while imaging artifacts and anatomical structures can produce similar responses. To address this problem, we propose STUNet-Fusion, a spatiotemporal framework for needle-tip localization in ultrasound videos. The proposed method formulates the input as a tri-channel spatio-temporal fusion tensor, comprising grayscale appearance, grid-based motion feature, and raw frame difference. A shared ResNet-34 encoder extracts spatial features, ConvLSTM integrates temporal dependencies, and a U-Net decoder reconstructs a dense probability heatmap. The final coordinates are extracted via a soft-argmax operation to achieve sub-pixel localization accuracy. Experimental results demonstrate that this spatiotemporal fusion strategy significantly improves localization robustness compared to conventional baselines.
\end{abstract}


\section{Introduction}
\label{sec:introduction}

Ultrasound-guided needle procedures are widely used in clinical interventions
because ultrasound provides real-time imaging without ionizing radiation.
Accurate localization of the needle tip is essential for safe and precise
needle placement, especially when the target region is small or close to
sensitive anatomical structures.

However, automatic needle-tip localization in ultrasound remains challenging.
Needle visibility is affected by insonation angle, imaging-plane mismatch,
insertion depth, and surrounding tissue. The needle tip may appear weak,
discontinuous, or partially invisible, while speckle, reverberation, shadowing,
and line-like anatomical structures can produce needle-like responses. These
factors make it difficult to distinguish the true needle tip from background
artifacts.

Many existing learning-based methods process each ultrasound frame
independently. Single-frame convolutional models have been used for needle
detection, needle-tip localization, and coordinate regression
\cite{mwikirize2018cnn,mwikirize2019single,groves2019localisation}. Although
these methods can capture spatial appearance information, they mainly rely on
static image information and do not explicitly use temporal evidence from
consecutive frames. As a result, localization can become ambiguous when the
needle tip has low contrast, is partially obscured, or resembles surrounding
structures.

Temporal information provides an important way to resolve this ambiguity \cite{ayvali2015optical, yan2023learning, che2024improving}.
During needle insertion, needle motion produces localized intensity changes
across consecutive ultrasound frames. Prior studies have explored digital
subtraction, time-aware neural networks, video-based learning, and motion-aware
segmentation to exploit temporal or motion information for ultrasound needle
analysis
\cite{mwikirize2019subtraction,mwikirize2021timeaware,rubin2021video,goel2024motionaware}.
These studies suggest that appearance and motion information are complementary
for improving needle localization robustness.

Based on this observation, this work proposes STUNet-Fusion, a spatiotemporal
heatmap-based framework for needle-tip localization in ultrasound video. The key
idea is to combine static appearance with motion information from consecutive
frames, so that the model can use both the visual structure of the needle and
its temporal movement information.

In the proposed method, each ultrasound frame is represented using three input
channels: grayscale appearance, grid-based motion feature, and raw frame difference. A shared ResNet-34 encoder extracts spatial features from each frame,
ConvLSTM integrates temporal information across the frame sequence, and a U-Net
decoder with skip connections reconstructs a dense needle-tip heatmap. The final
needle-tip coordinate is obtained from the peak response of the predicted
heatmap.

The main contributions of this work are as follows:
\begin{itemize}
    \item A three-channel input representation that combines grayscale
    appearance, grid-based motion feature, and raw frame difference for
    ultrasound needle-tip localization.

    \item A spatiotemporal heatmap-based architecture integrating ResNet-34,
    ConvLSTM, and a U-Net decoder with skip connections.

    \item A video-level evaluation including U-Net baseline comparison and
    ablation studies of temporal fusion, skip connections, and motion inputs.
\end{itemize}

\section{Related Work}
\label{sec:related}

\subsection{Single-Frame Needle Localization}

Automatic needle localization in ultrasound is challenging because the needle
may appear weak, fragmented, or partially invisible under unfavorable
insonation angles, while speckle, reverberation, and line-like anatomical
structures can produce needle-like responses. Earlier learning-based methods
commonly formulated this task as single-frame detection or coordinate
regression \cite{mwikirize2019single}. A fully convolutional proposal network
combined with a region-based detector was used to identify needle candidates
and estimate the needle trajectory and tip location \cite{mwikirize2018cnn}.
Another approach directly regressed the reflection centroid of an out-of-plane
needle from a single ultrasound image \cite{groves2019localisation}.

Although these methods demonstrated the effectiveness of convolutional
representations for needle localization, they mainly relied on static image
appearance. To address this limitation, we incorporate temporal evidence from
consecutive ultrasound frames, which provides additional motion cues when the
needle tip is weak, ambiguous, or partially obscured in a single frame.

\subsection{Motion-Based and Multi-Task Ultrasound Methods}

Temporal intensity variation provides useful information when the needle tip is
weak or ambiguous in a single ultrasound frame. Digital subtraction has been
used to enhance subtle changes caused by needle motion before applying a learned
detector or regression model \cite{mwikirize2019subtraction, yan2023learning,goel2024motionaware}. Time-aware deep
neural networks have also used consecutive ultrasound frames to improve
needle-tip localization under low-visibility conditions
\cite{mwikirize2021timeaware}. Video-based deep learning methods have further
shown that temporal encoding can improve ultrasound-guided needle insertion
analysis compared with frame-independent spatial models \cite{rubin2021video}.
Other studies have jointly addressed needle segmentation, tip detection, and
visibility estimation using multi-task networks, modified U-Net architectures,
and acquisition-side beam steering \cite{gao2021robust}.

These methods show that motion information can complement static appearance and
reduce ambiguity caused by ultrasound artifacts. We build on this idea by
combining grayscale appearance with both grid-based motion feature and raw frame difference, allowing the model to use motion cues at different spatial
scales.

\subsection{Spatial--Temporal Deep Architectures}

Residual networks provide effective per-frame feature extraction and improve
the optimization of deep convolutional models \cite{he2016resnet}. ConvLSTM
replaces fully connected state transitions with convolutional operations,
allowing temporal information to be integrated while preserving spatial
structure \cite{shi2015convlstm}. U-Net decoders and encoder--decoder skip
connections restore fine spatial details required for precise localization
\cite{ronneberger2015unet}.

Building on these components, we use a shared ResNet-34 encoder for per-frame
spatial feature extraction, ConvLSTM for temporal fusion across consecutive
frames, and a U-Net decoder with skip connections to reconstruct a dense
needle-tip heatmap.

\section{Methodology}
\label{sec:methodology}

\subsection{Overview}

\begin{figure*}[t]
    \centering
    \IfFileExists{STUNet_Fusion_Architecture.pdf}{%
        \includegraphics[width=0.98\textwidth]
        {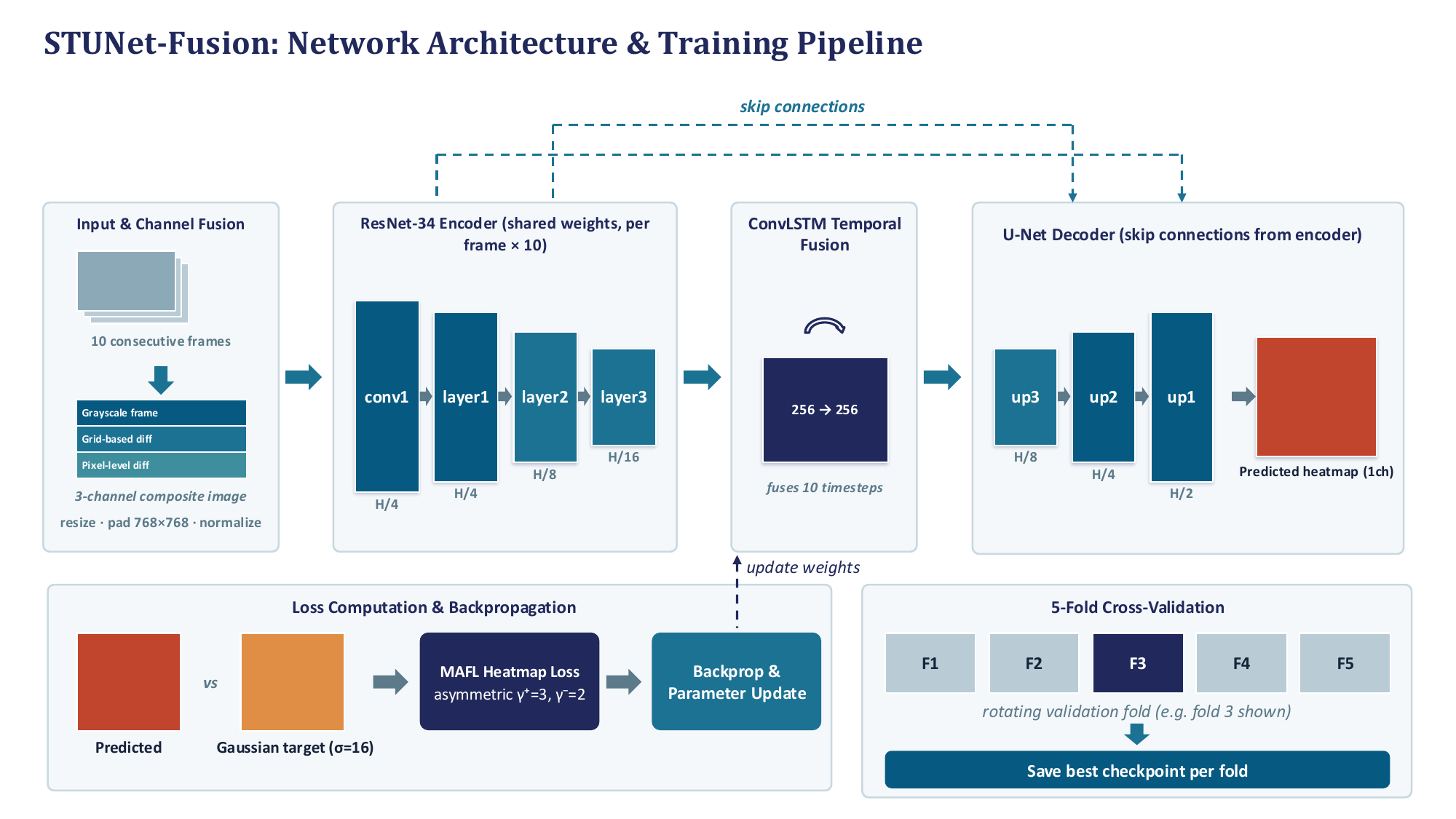}%
    }{%
        \fbox{\parbox[c][4.2cm][c]{0.94\textwidth}{\centering
        Upload \texttt{STUNet\_Fusion\_Architecture.pdf} to display the
        architecture figure.}}%
    }
    \caption{Overview of STUNet-Fusion. Ten three-channel fused frames are
    processed independently by a shared ResNet-34 encoder. The resulting
    feature sequence is integrated by ConvLSTM and decoded into a needle-tip
    heatmap using final-frame encoder skip connections.}
    \label{fig:architecture}
\end{figure*}

STUNet-Fusion localizes the needle tip from a short sequence of ultrasound
frames by combining appearance and motion information. As illustrated in
Fig.~\ref{fig:architecture}, the framework consists of four main stages:
three-channel input construction, per-frame spatial feature extraction using a
shared ResNet-34 encoder, temporal feature aggregation using ConvLSTM, and
heatmap reconstruction using a U-Net decoder. The model is supervised
with a Gaussian target heatmap and an asymmetric focal heatmap loss.

\subsection{Problem Formulation and Input Representation}

Let an ultrasound video clip contain $T$ consecutive frames
$\{I_t\}_{t=1}^{T}$ (where $I_t \in \mathbb{R}^{H\times W}$ denotes the $t$-th grayscale ultrasound image), where the objective is to localize the needle tip in the
final frame $I_T$. Rather than directly regressing an $(x,y)$ coordinate, the
model predicts a dense heatmap
$\hat{Y}\in\mathbb{R}^{H\times W}$. During training, the annotated needle-tip
coordinate in the final frame is converted into a Gaussian target heatmap
$Y\in\mathbb{R}^{H\times W}$. During inference, the predicted coordinate is
obtained via a soft-argmax operation over a local window $\Omega$ centered around the peak response of the sigmoid-normalized heatmap (i.e., $\sigma(\cdot)$) to achieve sub-pixel accuracy:

\begin{equation}
(\hat{x},\hat{y})
=
\frac{\sum_{(x,y) \in \Omega} (x,y) \cdot \sigma\!\left(\hat{Y}_{y,x}\right)}{\sum_{(x,y) \in \Omega} \sigma\!\left(\hat{Y}_{y,x}\right)},
\end{equation}

where $\sigma(\cdot)$ denotes the sigmoid function, and $\Omega$ represents a predefined local patch surrounding the maximum activation.

To represent both needle appearance and inter-frame motion, each frame is
converted to grayscale and paired with its immediately preceding frame. The
pixel-level motion image is computed as the absolute frame difference

\begin{equation}
D_t = \left|I_t-I_{t-1}\right|.
\end{equation}

A coarse regional-motion image $G_t$ is constructed from $D_t$. Specifically, the difference image $D_t$ is partitioned into non-overlapping blocks to form a $16 \times 16$ spatial grid. The local mean intensity is then computed for each grid cell. To suppress low-amplitude variations and imaging noise, a fixed threshold is applied at the grid level: if the mean intensity of a block is greater than the threshold, its value is retained; otherwise, it is suppressed to zero. Finally, the resulting $16 \times 16$ grid is resized back to the original image resolution using nearest-neighbor interpolation. This operation emphasizes regional changes while reducing the influence of small frame-to-frame intensity fluctuations.

The input representation for frame $t$ is defined as

\begin{equation}
F_t =
\operatorname{stack}\left(I_t,G_t,D_t\right),
\end{equation}
\hspace*{0em}where the three channels correspond to grayscale appearance, grid-based motion feature, and raw frame difference, respectively. A complete input
clip is therefore represented as

\begin{equation}
F =
\{F_t\}_{t=1}^{T}
\in
\mathbb{R}^{T\times3\times H\times W}.
\end{equation}

Each fused frame is resized while preserving its original aspect ratio and is
center-padded to $768\times768$ pixels. Pixel values are scaled to the range
$[0,1]$ and normalized using the ImageNet mean and standard deviation. The same
resize, scaling, and padding transformation is applied to the annotated
needle-tip coordinate. The implementation uses $T=10$ consecutive frames for
each input clip.

\subsection{Spatial Encoder and Temporal Fusion}

A ResNet-34 encoder with shared weights independently processes the $T$ fused
frames. The batch and temporal dimensions are first merged so that the same
encoder is applied to every frame:

\begin{equation}
F_{\mathrm{flat}}
\in
\mathbb{R}^{BT\times3\times H\times W},
\end{equation}
where $B$ denotes the batch size. The encoder consists of the initial
convolutional block followed by the first three residual stages. Features from
the third residual stage are restored to sequence form as

\begin{equation}
X
\in
\mathbb{R}^{B\times T\times256\times H/16\times W/16}.
\end{equation}

The sequence of deep feature maps is then processed by a ConvLSTM cell. Unlike
a conventional fully connected LSTM, ConvLSTM applies convolutional operations
within its recurrent gates and therefore preserves the two-dimensional spatial
layout. At time step $t$, the current encoder feature $X_t$ is combined with
the previous hidden state $H_{t-1}$ and cell state $C_{t-1}$. The recurrent
updates are

\begin{align}
i_t &= \sigma\!\left(W_i * [X_t,H_{t-1}] + b_i\right),\\
f_t &= \sigma\!\left(W_f * [X_t,H_{t-1}] + b_f\right),\\
o_t &= \sigma\!\left(W_o * [X_t,H_{t-1}] + b_o\right),\\
g_t &= \tanh\!\left(W_g * [X_t,H_{t-1}] + b_g\right),\\
C_t &= f_t \odot C_{t-1} + i_t \odot g_t,\\
H_t &= o_t \odot \tanh(C_t),
\end{align}
where $*$ denotes convolution, $\odot$ denotes element-wise multiplication,
and $[\cdot,\cdot]$ denotes channel-wise concatenation. After all $T$ frames
have been processed, the final hidden state

\begin{equation}
H_T
\in
\mathbb{R}^{B\times256\times H/16\times W/16}
\end{equation}

summarizes the accumulated spatial and temporal evidence across the clip.

\subsection{Heatmap Decoder and Skip Connections}

The final ConvLSTM hidden state $H_T$ is decoded through three upsampling stages. To restore high-resolution spatial information that may be weakened during deep encoding and temporal aggregation, we employ skip connections. Specifically, only the encoder features from the final frame $T$ are used in the skip pathways, while the ConvLSTM branch integrates information from all $T$ frames. Let $X^{(1)}_T$ and $X^{(2)}_T$ denote the spatial feature maps extracted from the first (\texttt{layer1}) and second (\texttt{layer2}) residual stages of the ResNet-34 encoder for the target frame $T$, respectively.

In the first decoding stage, the transposed convolution $\operatorname{up}_3(\cdot)$ increases the spatial resolution of $H_T$ from $H/16$ to $H/8$ and reduces the number of channels from 256 to 128. Its output is concatenated with $X^{(2)}_T$:
\begin{equation}
D_3 =
\phi_3\!\left(
\operatorname{concat}
\left(
\operatorname{up}_3(H_T),
X^{(2)}_T
\right)
\right),
\end{equation}
where $\operatorname{concat}(\cdot, \cdot)$ represents channel-wise concatenation, and $\phi_3(\cdot)$ is a composite function consisting of a $3\times3$ convolution, batch normalization, and ReLU activation.

The second transposed convolution $\operatorname{up}_2(\cdot)$ increases the resolution from $H/8$ to $H/4$ and reduces the feature dimension from 128 to 64. Its output is concatenated with $X^{(1)}_T$:
\begin{equation}
D_2 =
\phi_2\!\left(
\operatorname{concat}
\left(
\operatorname{up}_2(D_3),
X^{(1)}_T
\right)
\right),
\end{equation}
where $\phi_2(\cdot)$ denotes the corresponding composite convolution operation.

Finally, a third transposed convolution $\operatorname{up}_1(\cdot)$ increases the feature resolution to $H/2$. A convolutional prediction head $\operatorname{Head}(\cdot)$ collapses the feature channels to generate a single-channel heatmap. The output is finally resized to the original input resolution using bilinear interpolation, denoted as $\operatorname{Interp}(\cdot)$:
\begin{equation}
\hat{Y}
=
\operatorname{Interp}
\left(
\operatorname{Head}
\left(
\operatorname{up}_1(D_2)
\right)
\right).
\end{equation}

\subsection{Gaussian Heatmap Supervision}

For an annotated needle-tip coordinate $(c_x,c_y)$ in the resized and padded
image, the target heatmap is defined as

\begin{equation}
Y_{y,x}
=
\exp\left(
-\frac{(x-c_x)^2+(y-c_y)^2}{2\sigma_h^2}
\right),
\end{equation}
where $\sigma_h=16$ pixels controls the spatial spread of the target. The
Gaussian representation assigns the maximum value of one to the annotated tip
location and gradually decreases the supervision strength with increasing
distance from the center.

Compared with a one-pixel target, Gaussian supervision provides a smoother
optimization landscape and offers tolerance to small annotation or prediction
deviations. It also encourages the network to produce a spatially coherent
response around the needle tip rather than an isolated activation.

\subsection{Asymmetric Focal Heatmap Loss}

The predicted heatmap is optimized using an asymmetric focal loss designed for
the strong imbalance between the needle-tip region and the ultrasound
background. Let $p=\sigma(\hat{Y})$ denote the predicted probability map. The
positive and negative loss terms are

\begin{align}
\L_{\mathrm{pos}}
&=
(1-\alpha)(1-p)^{\gamma_{+}}\log(p),\\
\L_{\mathrm{neg}}
&=
\alpha p^{\gamma_{-}}(1-Y)^{\beta}\log(1-p).
\end{align}

The Gaussian-dependent term $(1-Y)^{\beta}$ reduces the penalty for negative
pixels near the annotated tip, while the asymmetric focusing terms emphasize
missed target responses and suppress hard background activations. The total
loss is defined as

\begin{equation}
\L_{\mathrm{total}}
=
-\frac{
\sum \L_{\mathrm{pos}}+\sum \L_{\mathrm{neg}}
}{
N_{\mathrm{pos}}HW
},
\end{equation}
where $N_{\mathrm{pos}}$ denotes the number of positive locations. The loss is
normalized by both the positive count and the number of spatial pixels.

\section{Experiments}
\label{sec:experiments}

\subsection{Experimental Setting}

\begin{figure}[t]
    \centering
    \includegraphics[width=0.32\linewidth]{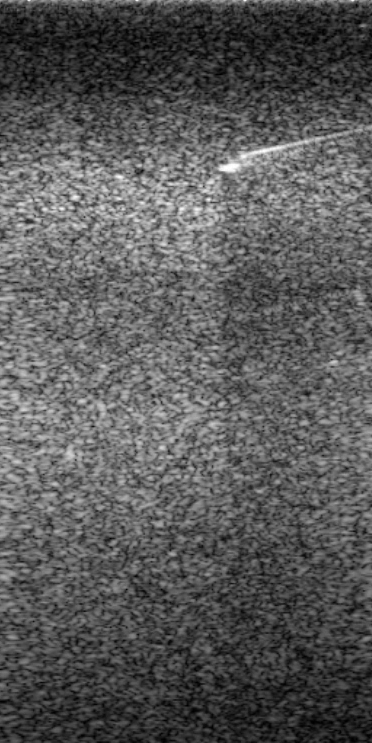}\hfill
\includegraphics[width=0.32\linewidth]{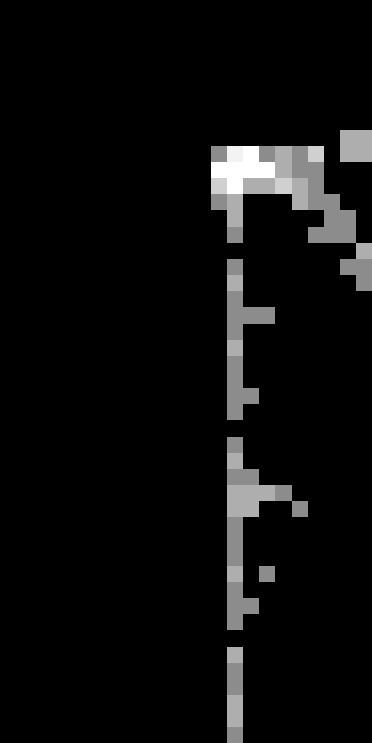}\hfill
\includegraphics[width=0.32\linewidth]{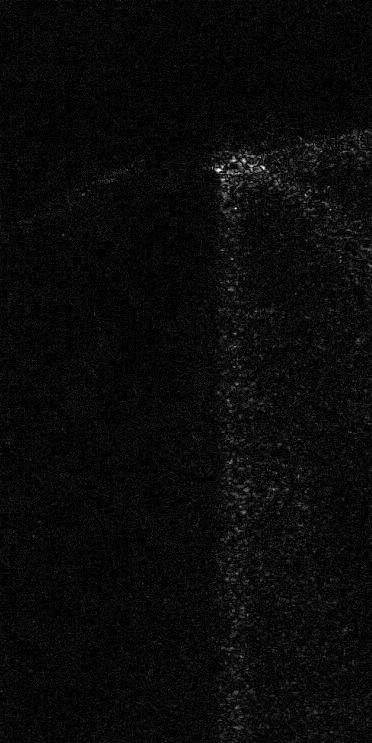}
    \caption{Example of the three input channels used in STUNet-Fusion.
From left to right: grayscale appearance $I_t$, grid-based motion feature $G_t$,
and raw frame difference $D_t$.}
    \label{fig:input}
\end{figure}

The experimental setup is organized into two main aspects: the dataset used for training and evaluation, and the implementation details of the proposed STUNet-Fusion framework.

\textbf{Datasets.} We collected the ultrasound needle dataset by recording needle insertion videos on a tissue-mimicking phantom using an ultrasound imaging system and Prodigy ultrasound imaging system (S-Sharp, New Taipei, Taiwan). Examples of the three input channels used by STUNet-Fusion are shown in Fig.~\ref{fig:input}. Each frame was manually annotated with the needle-tip location as a two-dimensional Cartesian coordinate $(x, y)$. The coordinate
annotations were used as ground-truth labels and converted into Gaussian
target heatmaps for heatmap-based training. The dataset was evaluated using
video-level five-fold cross-validation to avoid frame-level leakage between
training and validation sets. 

\textbf{Implementation Details.} We use video-level five-fold cross-validation with a fixed random seed. For
each fold, the model is trained for 70 epochs using Adam with a learning rate
of $10^{-4}$ and a batch size of two. Each input contains $T=10$ frames resized
and padded to $768\times768$, and the ResNet-34 encoder is initialized with
ImageNet-pretrained weights. The Gaussian target uses $\sigma=16$, while the
loss parameters are set to $\gamma_{+}=3$, $\gamma_{-}=2$, $\alpha=0.1$, and
$\beta=4$. The checkpoint with the lowest validation loss is retained.

\begin{table*}[t]
\centering
\caption{Best observed video-level performance of the baseline methods and
the proposed STUNet-Fusion model.}
\label{tab:baseline_best}
\begin{tabular}{lcccc}
\hline
\textbf{Method} &
\textbf{MLE (mm) $\downarrow$} &
\textbf{MedLE (mm) $\downarrow$} &
\textbf{RMSE (mm) $\downarrow$} &
\textbf{SDR@2\,mm (\%) $\uparrow$} \\
\hline

U-Net \cite{ronneberger2015unet}
& 44.50
& 8.92
& 68.62
& 13.4 \\

Time-aware DNN \cite{mwikirize2021timeaware}
& 41.17
& 3.52
& 58.14
& 26.7 \\

\textbf{STUNet-Fusion}
& \textbf{0.86}
& \textbf{0.81}
& \textbf{0.99}
& \textbf{98.2} \\

\hline
\end{tabular}
\end{table*}

\begin{table*}[t]
\centering
\caption{Mean $\pm$ standard deviation of the baseline comparison results.}
\label{tab:baseline_mean_std}
\begin{tabular}{lcccc}
\hline
\textbf{Method} &
\textbf{MLE (mm) $\downarrow$} &
\textbf{MedLE (mm) $\downarrow$} &
\textbf{RMSE (mm) $\downarrow$} &
\textbf{SDR@2\,mm (\%) $\uparrow$} \\
\hline

U-Net \cite{ronneberger2015unet}
& $57.00 \pm 6.84$
& $30.00 \pm 36.92$
& $79.20 \pm 7.54$
& $1.84 \pm 3.97$ \\

Time-aware DNN \cite{mwikirize2021timeaware}
& $55.38 \pm 8.87$
& $26.60 \pm 38.99$
& $79.15 \pm 11.46$
& $11.50 \pm 8.03$ \\

\textbf{STUNet-Fusion}
& $\mathbf{5.26 \pm 3.02}$
& $\mathbf{1.25 \pm 0.25}$
& $\mathbf{19.99 \pm 11.34}$
& $\mathbf{76.38 \pm 10.81}$ \\

\hline
\end{tabular}
\end{table*}

\subsection{Evaluation Metrics.}
The localization performance is evaluated using four metrics commonly adopted in ultrasound needle-tip localization studies \cite{mwikirize2021timeaware}: mean localization error (MLE), median localization error (MedLE), root mean squared error (RMSE), and successful detection rate within 2 mm (SDR@2 mm). For each test frame, the
predicted needle-tip coordinate $(\hat{x}_i, \hat{y}_i)$ is obtained from the
maximum response of the predicted heatmap and compared with the ground-truth
coordinate $(x_i, y_i)$. The localization error for frame $i$ is computed as
the Euclidean distance between the predicted and annotated tip locations:
\begin{equation}
e_i = \sqrt{(\hat{x}_i-x_i)^2 + (\hat{y}_i-y_i)^2}.
\end{equation}

Given $N$ evaluated frames, MLE measures the average localization error:
\begin{equation}
\mathrm{MLE} = \frac{1}{N}\sum_{i=1}^{N} e_i .
\end{equation}

MedLE measures the median localization error:
\begin{equation}
\mathrm{MedLE} = \mathrm{median}\left(\{e_i\}_{i=1}^{N}\right).
\end{equation}

RMSE gives larger penalty to large localization errors:
\begin{equation}
\mathrm{RMSE} = \sqrt{\frac{1}{N}\sum_{i=1}^{N} e_i^2}.
\end{equation}

SDR@2 mm measures the percentage of predictions whose localization error is
within 2 mm of the ground-truth needle-tip position:
\begin{equation}
\mathrm{SDR@2mm} =
\frac{1}{N}\sum_{i=1}^{N} \mathbb{I}(e_i \leq 2~\mathrm{mm}) \times 100\%,
\end{equation}
where $\mathbb{I}(\cdot)$ is the indicator function. Lower MLE, MedLE, and RMSE
indicate better localization accuracy, whereas a higher SDR@2 mm indicates a
larger proportion of accurately localized needle tips. Table~\ref{tab:baseline_mean_std}
reports the mean and standard deviation of the baseline comparison results.

\subsection{Comparison with Baselines}

The proposed STUNet-Fusion model was compared with two baseline methods:
a conventional U-Net and the time-aware deep neural network proposed by
Mwikirize et al.~\cite{ronneberger2015unet,mwikirize2021timeaware}.

\textbf{U-Net baseline.}
U-Net is a widely used encoder--decoder architecture for biomedical image
segmentation and localization, consisting of a contracting path for feature
extraction and an expanding path with skip connections for recovering spatial
details~\cite{ronneberger2015unet}. In our comparison, U-Net processes a
single grayscale ultrasound frame and predicts a single-channel needle-tip
heatmap. This baseline therefore evaluates needle-tip localization using only
spatial appearance information without explicit temporal modeling or
motion-based input channels.

\textbf{Time-aware DNN baseline.}
The time-aware deep neural network was developed specifically for
needle-tip localization in 2D ultrasound~\cite{mwikirize2021timeaware}.
The method enhances needle-tip motion across consecutive ultrasound frames
and combines convolutional feature extraction with LSTM-based temporal
modeling. It therefore provides a temporal baseline for evaluating whether
the proposed spatiotemporal fusion strategy offers additional benefit beyond
conventional recurrent modeling.

The same localization metrics were used for comparison. Lower MLE, MedLE, and
RMSE values indicate better localization performance, whereas a higher
SDR@2\,mm indicates a larger proportion of predictions located within 2\,mm
of the annotated needle-tip position.

Table~\ref{tab:baseline_best} reports the best observed video-level performance
for each method, while Table~\ref{tab:baseline_mean_std} reports the mean and
standard deviation across the evaluated videos. As shown in both tables,
STUNet-Fusion achieved lower localization errors and a higher SDR@2\,mm than
both the U-Net and Time-aware DNN baselines, indicating improved overall
needle-tip localization performance.

\begin{table}[t]
\centering
\caption{Best video-level performance of each ablation variant.}
\label{tab:ablation_best}
\resizebox{\columnwidth}{!}{
\begin{tabular}{lcccc}
\hline
\textbf{Model} &
\textbf{MLE (mm) $\downarrow$} &
\textbf{MedLE (mm) $\downarrow$} &
\textbf{RMSE (mm) $\downarrow$} &
\textbf{SDR@2\,mm (\%) $\uparrow$} \\
\hline
No Skip Connection
& 4.70
& 1.07
& 21.35
& 80.5 \\

No ConvLSTM
& 61.54
& 27.92
& 87.84
& 33.5 \\

Appearance Only
& 2.66
& 1.16
& 8.87
& 82.1 \\

\textbf{Full Model (STUNet-Fusion)}
& \textbf{0.86}
& \textbf{0.81}
& \textbf{0.99}
& \textbf{98.2} \\
\hline
\end{tabular}
}
\end{table}

\begin{table}[t]
\centering
\caption{Mean $\pm$ standard deviation of the ablation results averaged over
15 test videos.}
\label{tab:ablation_mean_std}
\resizebox{\columnwidth}{!}{
\begin{tabular}{lcccc}
\hline
\textbf{Model} &
\textbf{MLE (mm) $\downarrow$} &
\textbf{MedLE (mm) $\downarrow$} &
\textbf{RMSE (mm) $\downarrow$} &
\textbf{SDR@2\,mm (\%) $\uparrow$} \\
\hline
No Skip Connection
& $19.95 \pm 10.32$
& $1.73 \pm 0.50$
& $47.68 \pm 15.07$
& $58.48 \pm 16.48$ \\

No ConvLSTM
& $74.53 \pm 9.45$
& $60.26 \pm 31.92$
& $100.24 \pm 8.44$
& $24.90 \pm 6.50$ \\

Appearance Only
& $28.37 \pm 33.59$
& $15.01 \pm 33.72$
& $46.53 \pm 31.86$
& $40.92 \pm 26.19$ \\

\textbf{Full Model (STUNet-Fusion)}
& $\mathbf{5.26 \pm 3.02}$
& $\mathbf{1.25 \pm 0.25}$
& $\mathbf{19.99 \pm 11.34}$
& $\mathbf{76.38 \pm 10.81}$ \\
\hline
\end{tabular}
}
\end{table}

\begin{figure*}[t]
    \centering
    \includegraphics[width=0.32\linewidth]{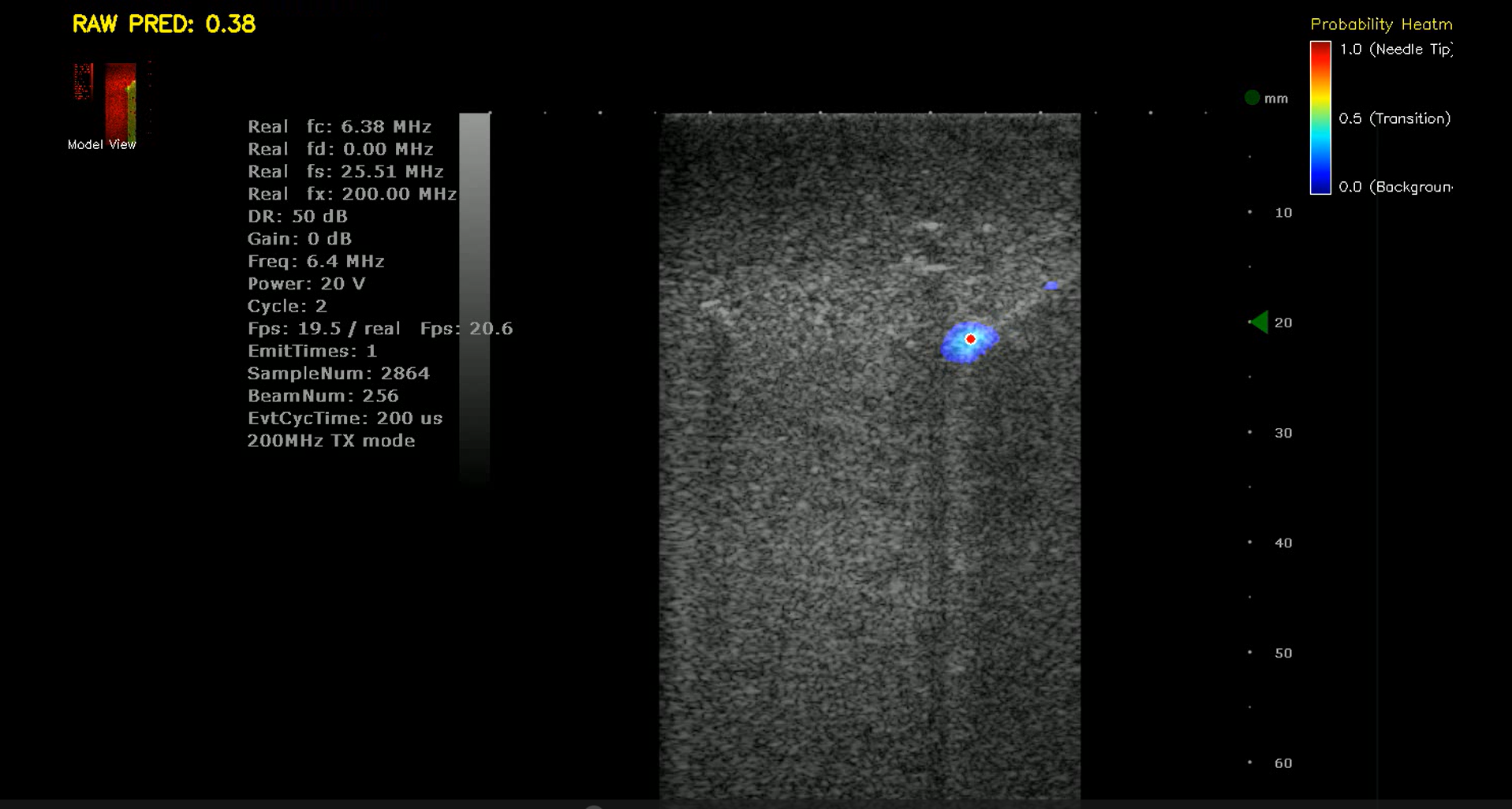}
    \hfill
    \includegraphics[width=0.32\linewidth]{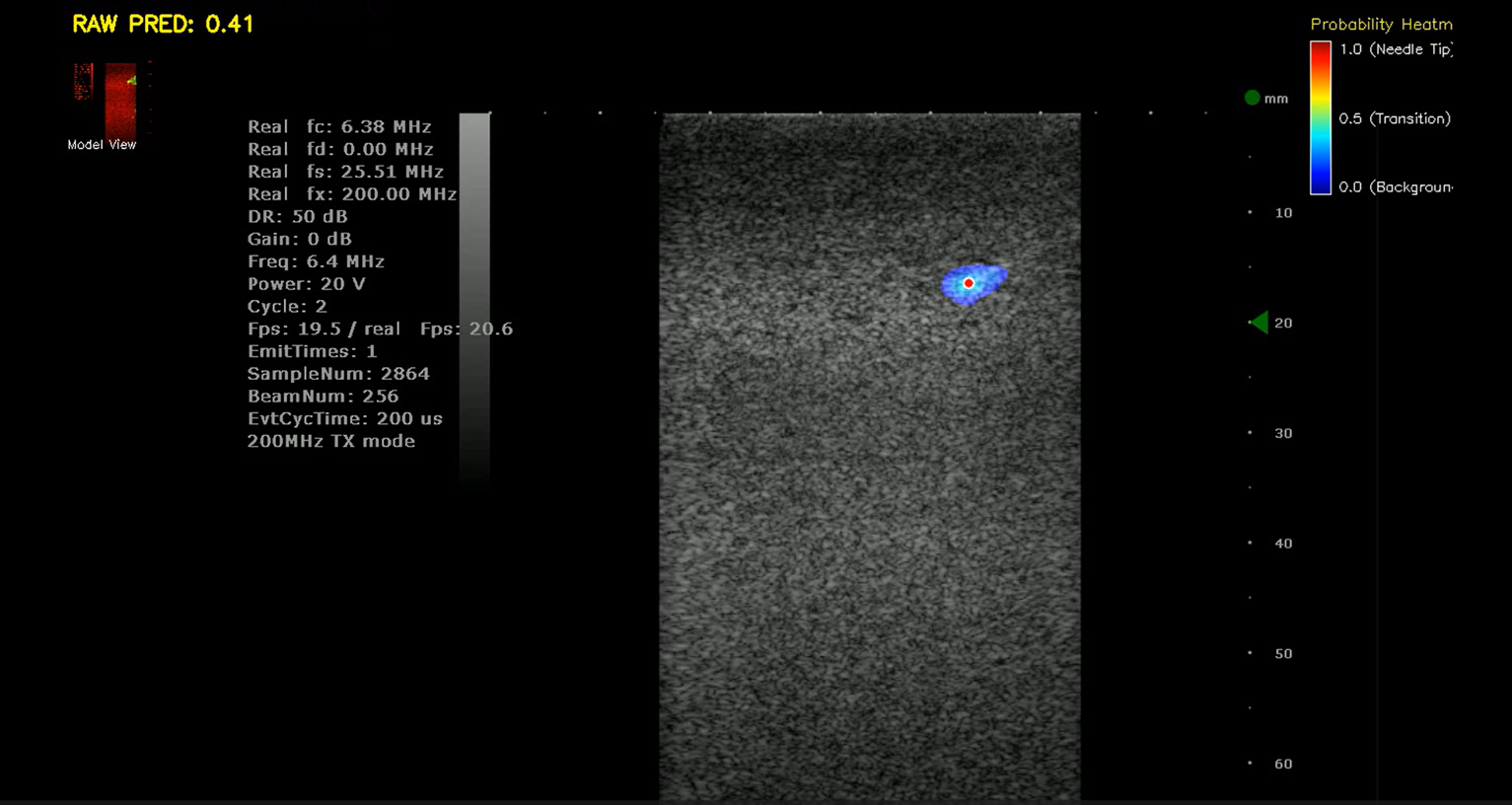}
    \hfill
    \includegraphics[width=0.32\linewidth]{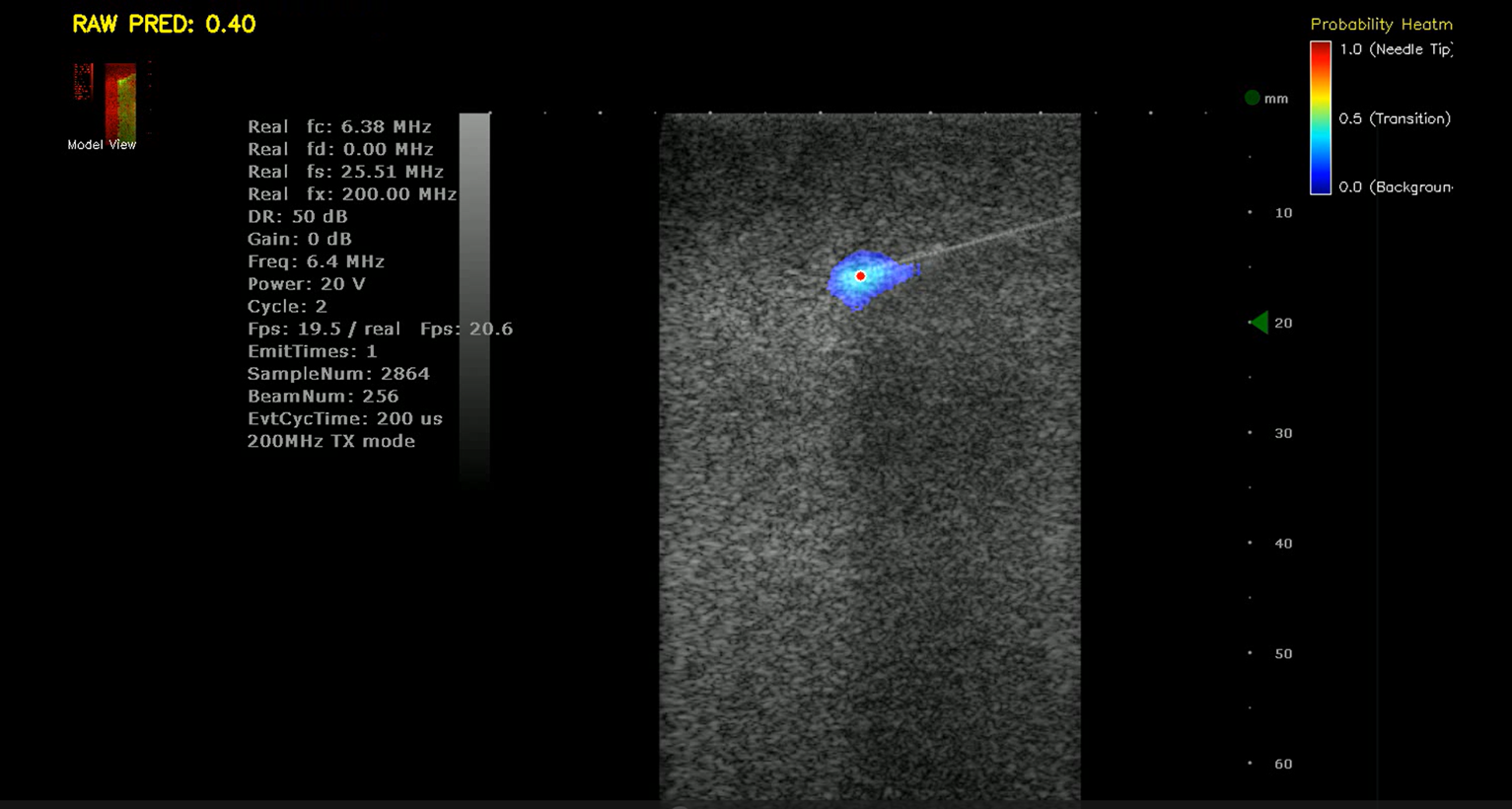}
    \caption{Representative needle-tip localization results. Predicted heatmaps are overlaid on the ultrasound frames to indicate the localized tip positions.}
    \label{fig:output_visualization}
\end{figure*}

\subsection{Ablation Study}

Since no external baseline model is currently evaluated under the same dataset
and experimental protocol, the comparison is presented through an ablation
study. The ablation experiments examine the contribution of the three major
components of STUNet-Fusion: the encoder--decoder skip connections, the
ConvLSTM-based temporal fusion module, and the appearance--motion fused input.

Two types of results are reported. Table~\ref{tab:ablation_best} presents the
best video-level result obtained by each model variant, while
Table~\ref{tab:ablation_mean_std} reports the mean and standard deviation
averaged over 15 test videos. For MLE, MedLE, and RMSE, lower values indicate
better localization performance. For SDR@2\,mm, a higher value indicates that a
larger proportion of predictions are located within 2\,mm of the annotated
needle-tip position.

\textbf{Effect of temporal fusion.}
Removing the ConvLSTM caused the largest performance degradation among all
ablation variants. The mean MLE increased from
$5.26 \pm 3.02$\,mm to $74.53 \pm 9.45$\,mm, while the mean SDR@2\,mm
decreased from $76.38 \pm 10.81\%$ to $24.90 \pm 6.50\%$.
The MedLE and RMSE also increased substantially. These results indicate that
temporal information from consecutive ultrasound frames is essential for
distinguishing needle-tip motion from background structures and imaging noise.

\textbf{Effect of skip connections.}
Removing the encoder--decoder skip connections increased the mean MLE from
$5.26 \pm 3.02$\,mm to $19.95 \pm 10.32$\,mm and the RMSE from
$19.99 \pm 11.34$\,mm to $47.68 \pm 15.07$\,mm. However, the MedLE remained
relatively low at $1.73 \pm 0.50$\,mm. This difference between the MedLE and
RMSE suggests that the model still localized the needle tip accurately in many
frames, but produced several large localization errors. The skip connections
therefore appear to be important for preserving high-resolution spatial details
during heatmap reconstruction and reducing severe localization failures.

\textbf{Effect of appearance--motion fusion.}
The Appearance Only variant used the grayscale ultrasound frame without the
grid-based motion feature channel $G_t$ or the raw frame difference channel
$D_t$. Its mean MLE increased to $28.37 \pm 33.59$\,mm, and its SDR@2\,mm
decreased to $40.92 \pm 26.19\%$. The large standard deviations indicate that
the performance of this variant varied considerably across videos. These
results suggest that appearance information alone is insufficient for stable
needle-tip localization and that the motion channels provide useful cues for
identifying the moving needle tip in noisy ultrasound images.

\textbf{Overall performance.}
The full STUNet-Fusion model achieved the best overall result for all four
evaluation metrics. It obtained an MLE of $5.26 \pm 3.02$\,mm, a MedLE of
$1.25 \pm 0.25$\,mm, an RMSE of $19.99 \pm 11.34$\,mm, and an SDR@2\,mm of
$76.38 \pm 10.81\%$. In the best-performing video, the model further achieved
an MLE of $0.86$\,mm, a MedLE of $0.81$\,mm, an RMSE of $0.99$\,mm, and an
SDR@2\,mm of $98.2\%$. Overall, the results show that temporal fusion,
high-resolution skip connections, and appearance--motion input fusion each
contribute to the final localization performance.

\subsection{Qualitative Results}
Fig.~\ref{fig:output_visualization} shows representative localization outputs. The predicted heatmaps, overlaid on the ultrasound frames, highlight a compact high-response region at the true needle tip. These examples demonstrate that STUNet-Fusion effectively focuses on the target and filters out speckle noise, even when the needle appearance is weak.

\section{Conclusion}
\label{sec:conclusion}
This paper presented STUNet-Fusion, a spatiotemporal framework for needle-tip
localization in ultrasound video. The proposed method combines grayscale
appearance and motion information from consecutive frames, using a shared
ResNet-34 encoder, ConvLSTM temporal fusion, and a U-Net decoder to predict a
dense needle-tip heatmap. Experimental results showed that STUNet-Fusion
outperformed the evaluated baseline methods, while the ablation study confirmed
the contributions of temporal fusion, skip connections, and motion-based input
information. These results demonstrate the effectiveness of combining spatial
and temporal information for robust ultrasound needle-tip localization.

\section{Limitations}
\label{sec:limitations}
This study has several limitations. First, the dataset was collected using a
tissue-mimicking phantom, so the model has not yet been fully validated on
in-vivo clinical ultrasound data. Second, the current experiments evaluate a
limited number of baseline methods, and additional comparisons with more recent
needle localization models are needed. Third, the model was evaluated in an
offline setting; future work should investigate real-time performance and
robustness during live ultrasound-guided needle procedures.

\section{Acknowledgments}
\label{sec:acknowledgments}
This work was supported by the National Science and Technology Council (NSTC),
Taiwan, under the Undergraduate Research Project program. The authors would also
like to thank Kai-Wei Lin and Bo-Ying Wang from the National Taiwan University
of Science and Technology for their support and assistance.

\bibliographystyle{IEEEbib}
\bibliography{refs}

\end{document}